\documentclass{llncs}
\usepackage[a4paper,includeheadfoot,top=1.65in,bottom=1.65in,left=1.73in,hcentering]{geometry}

\usepackage[utf8]{inputenc}
\usepackage{lmodern}
\usepackage{booktabs}
\usepackage{tabularx}
\usepackage{graphicx}
\usepackage{amsmath}

\usepackage[english]{babel}

\usepackage{xspace}
\newcommand{\sopa}{SoPa\xspace}
\newcommand{\sopaseq}{SoPa Seq2seq\xspace}
\newcommand{\lstmseq}{LSTM Seq2seq\xspace}
\newcommand{\unimorph}{UniMorph\xspace}
\newcommand{\langno}{29\xspace}
\newcommand{\goodlangno}{12\xspace}
\newcommand{\morphana}{morphological analysis\xspace}

\newcommand{\lemmatization}{lemmatization\xspace}
\newcommand{\Lemmatization}{Lemmatization\xspace}
\newcommand{\copytask}{copy\xspace}

\newcommand{\tabref}[1]{Table~\ref{tab:#1}}
\newcommand{\secref}[1]{Section~\ref{sec:#1}}
\newcommand{\figref}[1]{Fig.~\ref{fig:#1}}
\newcommand{\equref}[1]{Eq.~\ref{eq:#1}}

\usepackage{xcolor}
\newif{\ifhidecomments}
\hidecommentsfalse

\ifhidecomments
    \newcommand{\kornai}[1]{}
    \newcommand{\judit}[1]{}
\else
    \newcommand{\kornai}[1]{\textcolor{blue}{[#1 ({\bf Andras})]}} 
    \newcommand{\judit}[1]{\textcolor{brown}{[#1 ({\bf Judit})]}} 
\fi
\newif{\ifhidetodo}
\hidetodofalse

\ifhidetodo
    \newcommand{\todo}[1]{}
\else
    \newcommand{\todo}[1]{\textcolor{magenta}{[TODO: #1 ]}}
\fi

\begin{document}
\sloppy

\title{The Role of Interpretable Patterns in Deep Learning for Morphology}
\author{Judit \'Acs$^{1,2}$, Andr\'as Kornai$^2$}

\institute{
    $^1$ Department of Automation and Applied Informatics\\
    Budapest University of Technology and Economics\break
    $^2$ Institute for Computer Science and Control\\
}

\maketitle

\begin{abstract}

We examine the role of character patterns in three tasks: morphological
analysis, lemmatization and copy. We use a modified version of the standard
sequence-to-sequence model, where the encoder is a pattern matching network.
Each pattern scores all possible N character long subwords (substrings) on the
source side, and the highest scoring subword's score is used to initialize the
decoder as well as the input to the attention mechanism.  This method allows
learning which subwords of the input are important for generating the
output. By training the models on the same source but different target, we can
compare what subwords are important for different tasks and how they relate to
each other. We define a similarity metric, a generalized form of the Jaccard
similarity, and assign a similarity score to each pair of the three tasks that
work on the same source but may differ in target. We examine how these three
tasks are related to each other in \goodlangno languages. Our code is publicly
available.\footnote{https://github.com/juditacs/deep-morphology}

\end{abstract}

\section{Introduction}

Deep neural networks are successful at various morphological tasks as
exemplified in the yearly SIGMORPHON Shared
Task\cite{Cotterell:2016,Cotterell:2017,Cotterell:2018}. However these
neural networks operate with continuous representations and weights which is in
stark contrast with traditional, and hugely successful, rule-based morphology.
There have been attempts to add rule-based and discrete elements to these
models through various inductive biases\cite{Aharoni:2016}. 

In this paper we tackle two morphological tasks and the copy task as a control
with an interpretable model, \sopa.  Soft Patterns\cite{Schwartz:2018} or \sopa
is a finite-state machine parameterized with a neural network, that learns
linear patterns of predefined size. The patterns may contain epsilon
transitions and self-loops but otherwise are linear. \emph{Soft} refers to the
fact that the patterns are intended to learn abstract representations that may
have multiple surface representations, which \sopa can learn in an end-to-end
fashion. We call these surface representations \emph{subwords}, while the
abstract patterns, \emph{patterns} throughout the paper.  An important upside
of \sopa is that interpretable patterns can be extracted from each sample.
\cite{Schwartz:2018} shows that \sopa is able to retrieve meaningful word-level
patterns for sentiment analysis. Each pattern is matched against every possible
subword and the highest scoring subword is recovered via a differentiable
dynamic program, a variant of the forward algorithm.  We apply this model as
the encoder of a sequence-to-sequence or \emph{seq2seq}\footnote{also called
encoder-decoder model} model\cite{Sutskever:2014}, and add an
LSTM\cite{Hochreiter:1997} decoder.  We initialize the decoder's hidden state
with the final scores of each \sopa pattern and we also apply Luong's
attention\cite{Luong:2015b} on the intermediate outputs generated by \sopa. We
call this model \sopaseq.  We compare each setup to a sequence-to-sequence with
a bidirectional LSTM encoder, unidirectional LSTM decoder and Luong's
attention.

We show that \sopaseq is often competitive with the LSTM baseline while also
interpretable by design. \sopaseq is especially good at \morphana, often
surpassing the LSTM baseline, which confirm our linguistic intuition namely
that subword patterns are useful for extracting morphological information.

We also compare these models using a generalized form of Jaccard-similarity and
we find that some trends coincide with linguistic intuition.

\section{Data}

Universal Morphology or \unimorph is project that aims to improve how NLP
handles languages with complex
morphology.\footnote{\url{https://unimorph.github.io/}} Specified in
\cite{Sylak-Glassman:2016}, \unimorph has been used to annotate 350 languages
from the English edition of
Wiktionary\footnote{\url{https://en.wiktionary.org/}}. Wiktionary contains
inflection tables that list inflected forms of a word. Part of the \unimorph
project is converting these tables into \emph{(lemma, inflected form, tags)}
triplets such as \emph{(ablak, ablakban, N IN+ESS SG)}.  The first tag is the
part-of-speech which is limited to the main open classes (nouns, verbs and
adjectives) in most languages, \verb|IN+ESS| is the inessive case and \verb|SG|
denotes singular.

\subsection{Data sampling}

Our goal is to sample 10000 training, 2000 development and 2000 test examples.
We retrieved 109 \unimorph repositories (109 languages) but only 57 languages
have at least 14000 samples, the lowest possible number for our purposes.  We
first prefilter the languages and assign them to languages families and genus
using the World Atlas of Languages or WALS\footnote{\url{https://wals.info/}}.
WALS does not contain extinct, constructed or liturgical languages, and we do
not incorporate these in our dataset. Out of the 109 languages, 19 have no WALS
entry.  \langno languages have large enough \unimorph datasets that allow
obtaining 10000/2000/2000 samples.\footnote{Albanian has only 1982 test samples
but we wanted to include it as a language isolate from the Indo-European
family.}  \tabref{data_stats} summarizes the dataset.

\begin{table}[!t]
    \centering
{\scriptsize
\begin{tabular}{llllrrrrl}
\toprule
      Language &         Family &         Genus & sample &  lemma &  paradigm &  alphabet &    F/L &  POS \\
\midrule
        Arabic &   Afro-Asiatic &       Semitic &   138k &   4007 &       196 &        45 &   26.3 &  NVA \\
       Turkish &         Altaic &        Turkic &   213k &   3017 &       186 &        46 &   54.7 &  NVA \\
       Quechua &          Hokan &         Yuman &   178k &   1003 &       553 &        22 &  146.8 &  NVA \\
      Albanian &  Indo-European &      Albanian &    14k &    587 &        59 &        27 &   17.4 &   NV \\
      Armenian &  Indo-European &      Armenian &   259k &   6991 &       134 &        46 &   35.3 &  NVA \\
       Latvian &  Indo-European &        Baltic &   129k &   7238 &        78 &        34 &   10.3 &  NVA \\
    Lithuanian &  Indo-European &        Baltic &    33k &   1391 &       139 &        56 &   20.1 &  NVA \\
         Irish &  Indo-European &        Celtic &    45k &   7299 &        53 &        53 &    3.3 &  NVA \\
        Danish &  Indo-European &      Germanic &    25k &   3190 &        14 &        44 &    7.7 &   NV \\
        German &  Indo-European &      Germanic &   171k &  15032 &        37 &        63 &    4.5 &   NV \\
       English &  Indo-European &      Germanic &   115k &  22765 &         5 &        65 &    4.0 &    V \\
     Icelandic &  Indo-European &      Germanic &    76k &   4774 &        44 &        54 &   10.9 &   NV \\
         Greek &  Indo-European &         Greek &   147k &  11872 &       118 &        76 &    6.5 &  NVA \\
       Kurdish &  Indo-European &       Iranian &   203k &  14143 &       128 &        61 &   14.3 &  NVA \\
      Asturian &  Indo-European &       Romance &    29k &    436 &       223 &        32 &   49.5 &  NVA \\
       Catalan &  Indo-European &       Romance &    81k &   1547 &        53 &        35 &   40.6 &    V \\
        French &  Indo-European &       Romance &   358k &   7528 &        48 &        44 &   35.3 &    V \\
     Bulgarian &  Indo-European &        Slavic &    54k &   2413 &        95 &        31 &   18.9 &  NVA \\
         Czech &  Indo-European &        Slavic &   109k &   5113 &       147 &        62 &   10.0 &  NVA \\
     Slovenian &  Indo-European &        Slavic &    59k &   2533 &        94 &        56 &    8.9 &  NVA \\
      Georgian &     Kartvelian &    Kartvelian &    74k &   3777 &       109 &        33 &   17.5 &  NVA \\
        Adyghe &   NW Caucasian &  NW Caucasian &    20k &   1635 &        30 &        40 &   11.9 &   NA \\
          Zulu &    Niger-Congo &       Bantoid &    49k &    566 &       249 &        46 &   57.2 &  NVA \\
       Khaling &   Sino-Tibetan &   Mahakiranti &   156k &    591 &       432 &        32 &   91.5 &    V \\
      Estonian &         Uralic &        Finnic &    27k &    886 &        64 &        26 &   28.0 &   NV \\
       Finnish &         Uralic &        Finnic &     1M &  57165 &        97 &        50 &   27.1 &  NVA \\
         Livvi &         Uralic &        Finnic &    63k &  15295 &       104 &        55 &    4.0 &  NVA \\
 Northern Sami &         Uralic &         Saami &    62k &   2103 &        80 &        31 &   25.9 &  NVA \\
     Hungarian &         Uralic &         Ugric &   517k &  14883 &        93 &        53 &   34.1 &   NV \\
\bottomrule
\end{tabular}

}
    \caption{\label{tab:data_stats} Dataset statistics. The
languages are sorted by language family. F/L refers to the form-per-lemma
ratio. POS indicates which part of speech are present in the dataset out of the
nouns, verbs and adjectives.
}
\end{table}


\section{Tasks}\label{sec:tasks}

We train both kinds of seq2seq models on three tasks: morphological analysis
(abbreviated as \emph{\morphana}), \emph{\lemmatization}, and
\emph{\copytask} or autoencoder. The source sequence is the inflected form of
the word in all three tasks, while the target sequence is a list of
morphosyntactic tags for \morphana, the lemma for \lemmatization and the same
as the source side for \copytask. \tabref{examples} shows examples for the
three tasks.

Inflected words and lemmas are treated as a sequence of characters but tags are
treated as standalone symbols. We share the vocabulary and the embedding
between the source and target side when training for \copytask and
\lemmatization but we use separate vocabularies for \morphana.

\begin{table*}[!t]
\centering
\begin{tabularx}{\textwidth}{lllX}
    \toprule
    Language & Task & Source & Target \\
    \midrule
    Hungarian & \morphana & vásároljanak & V SBJV PRS INDF 3 PL \\
    Hungarian & \morphana & lepkékben & N IN+ESS PL \\
    English & \morphana & hugging & V V.PTCP PRS \\
    French & \morphana & désinstalleriez & V COND 2 PL \\
    \midrule
    Hungarian & \lemmatization & vásároljanak & vásárol \\
    Hungarian & \lemmatization & lepkékben & lepke \\
    English & \lemmatization & hugging & hug \\
    French & \lemmatization & désinstalleriez & désinstaller \\
    \midrule
    Hungarian & \copytask & vásároljanak & vásároljanak \\
    Hungarian & \copytask & lepkékben & lepkékben \\
    English & \copytask & hugging & hugging \\
    French & \copytask & désinstalleriez & désinstalleriez \\
    \bottomrule
\end{tabularx}
    \caption{\label{tab:examples}Dataset examples.}
\end{table*}

\section{Models}

We train two kinds of sequence-to-sequence models which only differ in the
choice of the encoder. Both models first pass the input through an embedding.
We train the embeddings from randomly initialized values and do not use
pretrained embeddings. We use character embeddings with 50 dimensions for
character inputs and outputs and tag embeddings with 20 dimensions for
morphological tags (only for \morphana). The embeddings are shared between the
encoder and decoder for \lemmatization and \copytask, since both the source and
the target sequences are characters.  The output of the source embedding is the
input to the encoder module which is a \sopa with 120 patterns in \sopaseq case
and a bidirectional LSTM in the baseline.  The decoder later attends on the
intermediary outputs of these modules. The final hidden state of the encoder
module is used to initialize the decoder. The decoder side of these models is
identical in both setups, an LSTM with Luong's attention.  All LSTMs have 64
hidden cells and a single layer.

The size of \sopa patterns (3, 4, and 5 in our case) define the number of
forward arcs that a pattern has. These may contain epsilon steps and self loops
but an epsilon or a self loops is always followed by a main transition
(consuming an actual symbol). This means that a 3 long pattern may contain one
epsilon and one main transition, two epsilons or two main transitions. Any main
transition may be preceded by a self loop. The pattern size includes the start
state and the end state. In our experiments we used 3, 4, and 5 long patterns,
40 patterns of each length.

Most of the training details are also identical. We train with batch size 64,
and we use early stopping if the development loss and accuracy stop improving
for 5 epochs. We maximize the number of epochs in 200 but this is never
reached. We save the best model based on development accuracy. We use the Adam
optimizer with 0.001 learning rate for all experiments.

\sopa is more difficult to train than LSTMs, so we decay the learning rate by
0.5 if the development loss does not decrease for 4 epochs.

\section{Model similarity}

We define a similarity metric between two \sopaseq models measured on datasets
that share their source side. The target side may differ. The three tasks
introduced in \secref{tasks}, all take inflected word forms as their source
sequence, which allows computing our similarity metric between each pair of
tasks.

\sopa works with a predefined number of patterns and tries matching each
pattern on any subword of the input with a particular length.  The highest
scoring subword is used in the final source representation. We take the highest
scoring $T=10$ patterns for each input and compare the subwords that resulted
in these scores. The metric is defined as the average similarity over the
dataset $D$:

\begin{equation}\label{eq:fullsim}
    \text{Sim}(M_1, M_2, D) = \frac{1}{|D|} \sum_{d \in D} S(M_1(d), M_2(d)),
\end{equation}

\noindent where $M_1$ and $M_2$ are the models, and $S$ is the similarity of
the two representations generated by the encoder side of the models on sample
$d$, defined as:

\begin{equation}\label{eq:sample}
    S(M_1(d), M_2(d)) = \frac{1}{2T} (
    \sum_{p_i \in P_1} \max_{p_j \in P_2} J(p_i, p_j) +
    \sum_{p_j \in P_2} \max_{p_i \in P_1} J(p_i, p_j)
    ),
\end{equation}

\noindent where $T$ is a predefined number of highest scoring patterns on that
sample (10 in our experiments), $P_1$ is the set of $T$ highest scoring
patterns of $M_1$, $P_2$ is the set of $T$ highest scoring patterns of $M_2$
and $J$ is the Jaccard similarity of two subwords defined as the proportion of
overlapping symbols by the union of all symbols. Jaccard similarity is 0 if
there is no overlap and is 1 when the subwords are the same. For each sample,
we first choose the highest scoring $T$ patterns from each model, we denote
these sets of patterns as $P_1$ and $P_2$. Then we find the subwords
corresponding to these patterns. We compute the pairwise Jaccard similarities
between every element of $P_1$ and $P_2$. Then for each pattern, we find the
most similar pattern from the other model. The average of these scores is the
similarity of the two models on that sample (see \equref{sample}) and the
average over all samples (see \equref{fullsim}) is the similarity of two models
on dataset $D$.  This metric is symmetric and it ranges from 0 to 1.
Table~\ref{table:sim_example} shows a small example of this similarity on the
word \emph{ablakban}.

\begin{table}[!t]
\centering
\begin{tabular}{l|r|r|r|r|r}
 & \underline{\textasciicircum ab}lakban\$ & \textasciicircum
abl\underline{akb}an\$ & \textasciicircum ablak\underline{ban}\$ 
& \textasciicircum ab\underline{lakb}an & Max\\
\hline
\textasciicircum ablak\underline{ban}\$ & 0 & 0.2 & 1 & 0.75 & 1 \\
\hline
\textasciicircum abla\underline{kba}n\$ & 0 & 0.5 & 0.5 & 0.75 & 0.75 \\
\hline
\textasciicircum ab\underline{lak}ban\$ & 0 & 0.5 & 0 & 0.167 & 0.5\\
\hline
\textasciicircum abla\underline{kban}\$ & 0 & 0.75 & 0.167 & 0.333 & 0.75 \\
\hline
Max & 0 & 0.75 & 1 & 0.75 & J=0.6875 \\
\end{tabular}
\caption{\label{table:sim_example} Simlarity (\equref{sample}) between two
models $M_1$ and $M_2$ on one sample using the 4 highest scoring subwords
($T=4$) with the subwords underlined. Rows correspond to the highest scoring
subwords from $M_1$ (ban, kba, lak, kban), while columns correspond to the
subwords from $M_2$ (\textasciicircum ab, akb, ban, lakb). A Jaccard similarity
matrix (with position information) is constructed. The final similarity is the
mean maximum of every row and every column of the matrix.}
\end{table}


\section{Results and analysis}

\begin{figure}[!t]
    \includegraphics[width=\linewidth]{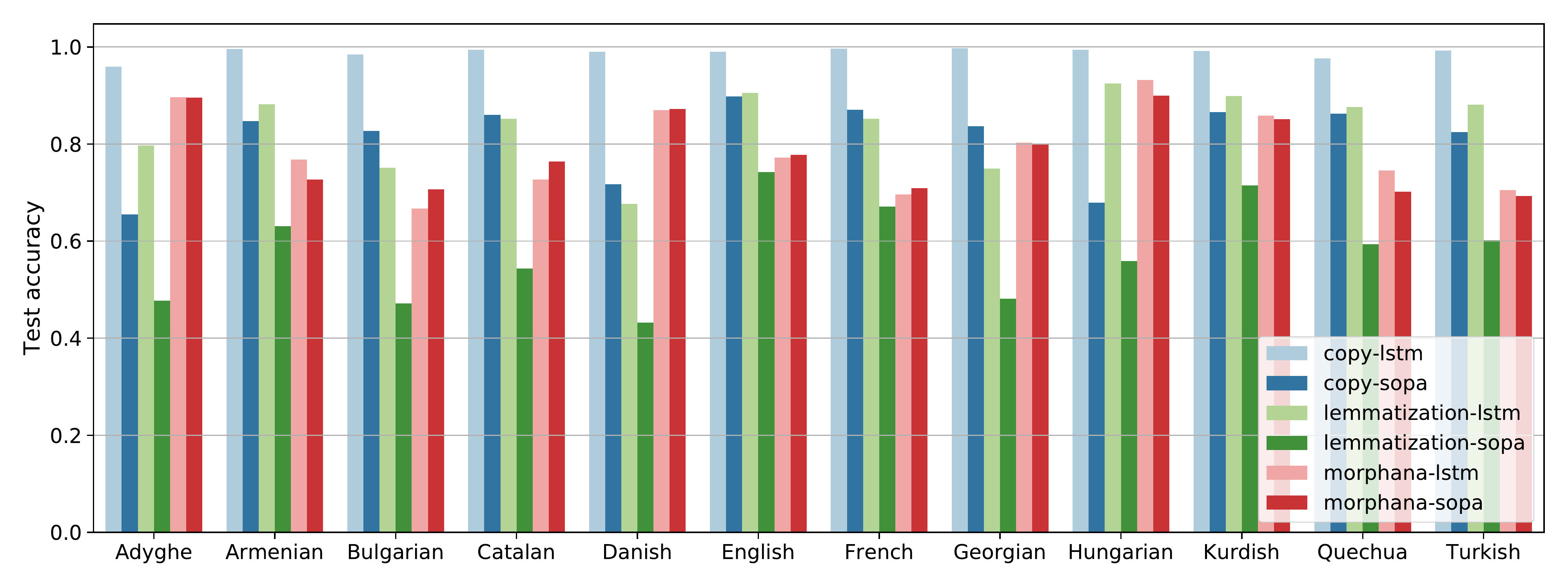}
    \caption{\label{fig:all_acc} Accuracy of \sopaseq models on each language
and task.}
\end{figure}

We first show that \sopaseq is competitive with the \lstmseq baseline,
especially for \morphana.  An output is considered accurate if it fully matches
the reference and we do not consider partial matching. Some languages prove to
be too difficult for the models, which may be due to the lack of context that
is often needed for \morphana and orthographic changes often present in
\lemmatization. We continue our analysis on languages where each of the three
tasks are performed by \sopa `reasonably well', which we set to 40\% accuracy
or higher on the development set. This leaves us with \goodlangno languages.
The reason we set a lower limit to accuracy is that we have no reason to
believe that a bad model's representation is useful for the task.
\figref{all_acc} shows the test accuracy in these languages. \Lemmatization is
consistently the most difficult task for \sopa, while \sopa is on pair with
\lstmseq in \morphana, sometimes outperforming it. We attribute this result to
the fact that a morphological tag often corresponds to a single morpheme,
usually with a few possible surface realizations that \sopa's `soft' patterns
can pick up on.  On the other hand \lemmatization and \copytask require
regenerating much of the input which is more difficult from an inherently
summarized representation such as the one \sopa generates.

\begin{figure}[!t]
    \includegraphics[width=\linewidth]{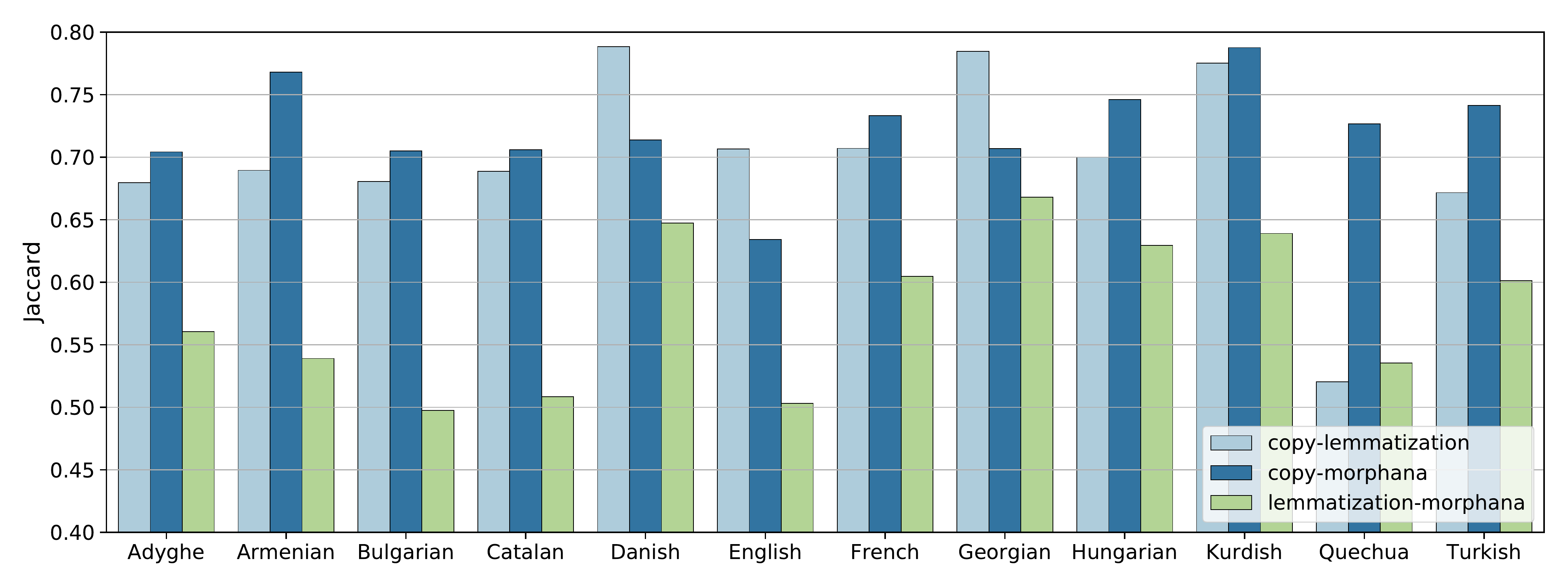}
    \caption{\label{fig:jaccard} Model similarity between all task pairs by
language. Higher similarity indicates that two models handle the same source in
a more similar way.}
\end{figure}

We continue by computing the pairwise similarity value defined in
\equref{fullsim} between the three tasks. Higher values indicate that \sopa
finds similar patterns valuable for generating the output. \figref{jaccard}
shows the pairwise similarity of models trained for the three tasks. We only
compute these similarities on samples where the output of \emph{both} models
are correct (generally 40-60\% of the test samples).

\Lemmatization and \morphana are the least similar in almost every language.
This is not surprising considering that \lemmatization is the task of discarding
information that \morphana needs to correctly tag.  Quechua is the only
exception from this trend which could be explained by the very rich
inflectional morphology (especially at the type-level) that results in lemmas
being significantly shorter than inflected forms. This means that \copytask
needs to memorize a lot more of the source word than \lemmatization.

Another trend we observe, is that \copytask and \morphana are more similar than
\copytask and \lemmatization in languages with rich inflectional morphology
such as Armenian, Hungarian, Kurdish and Turkish and the opposite is true in
fusional and morphologically poor languages such as Danish and English.
Georgian seems to be an exception and further exploration is an exciting
research direction.

Finally we demonstrate \sopa's interpretability by extracting the most
frequently matched subwords in each language and task. \tabref{top_subwords}
lists the most common subwords in English, French and Hungarian in each task.
It should be noted that these subwords are very short because we used 3, 4 and
5 long patterns that match 2, 3 and 4 characters not including self loops and
short patterns simply occur more frequently. 

\begin{table}[!t]
    \centering
\begin{tabular}{lll}
\toprule
  language &                    task &                           subwords \\
\midrule
   English &                    copy &     ed,e\$,ed\$,es,in,at,re,s\$,te,ri \\
   English &           lemmatization &     at,g\$,er,in,ng,iz,s\$,en,ize,es \\
   English &  morphological\_analysis &      d\$,s\$,e\$,es\$,\$,ed,ed\$,o,ng,g\$ \\
    French &                    copy &      s\$,ss,is,as,ie,ai,z\$,nt,ns,en \\
    French &           lemmatization &      er,s\$,t\$,nt,ie,ns,ra,is,ri,\textasciicircum d \\
    French &  morphological\_analysis &  s\$,t\$,z\$,nt\$,ez\$,e\$,ai,er,ns\$,es\$ \\
 Hungarian &                    copy &     l\$,n\$,k\$,sz,t\$,nk\$,kk,el,ok,na \\
 Hungarian &           lemmatization &      sz,t\$,k\$,l\$,ta,tá,\textasciicircum k,n\$,kb,ró \\
 Hungarian &  morphological\_analysis &      l\$,t\$,n\$,k\$,ek,a\$,\$,g\$,á\$,ak\$ \\
\bottomrule
\end{tabular}

    \caption{\label{tab:top_subwords} Top subwords extracted from English,
French and Hungarian. \textasciicircum and \$ denote word start and end
respectively. }
\end{table}

\section{Conclusion}

We presented an application of Soft Patterns -- a finite state automaton
parameterized by a neural network -- as the encoder of a sequence-to-sequence
model. We show that it is competitive with the popular LSTM encoder on
character-level copy and morphological tagging, while providing interpretable
patterns.

We analyzed the behavior of \sopa encoders on \morphana, \lemmatization and
\copytask by computing the average Jaccard similarity between the patterns
extracted from the source side. We found two trends that coincide with
linguistic intuition. One is that \lemmatization and morphological analysis
require patterns that match less similar subwords than the other two task
pairs. The other one is that \copytask and morphological analysis are more
similar in languages with rich inflectional morphology.

\section*{Acknowledgments}
Work partially supported by 2018-1.2.1-NKP-00008: Exploring the Mathematical
Foundations of Artificial Intelligence; and National Research, Development and
Innovation Office grant NKFIH \#120145 `Deep Learning of Morphological
Structure'. We thank Roy Schwartz for his help in understanding the inner
mechanics of \sopa.

\bibliographystyle{splncs}
\bibliography{ml}

\end{document}